\documentclass{article}

\usepackage{arxiv}

\usepackage[utf8]{inputenc} 
\usepackage[T1]{fontenc}    
\usepackage{hyperref}       
\usepackage{url}            
\usepackage{booktabs}       
\usepackage{amsfonts}       
\usepackage{nicefrac}       
\usepackage{microtype}      
\usepackage{lipsum}		
\usepackage{graphicx}
\usepackage{natbib}
\usepackage{doi}
\usepackage{amsmath}

\title{Combining Varied Learners for Binary Classification using Stacked Generalization}

\author{
  Sruthi Nair \\
  Master of Engineering \\
  Vidyalankar Institute of Technology \\
  Mumbai, India\\
  \texttt{sruthi.rk.nair@gmail.com} \\
  \And
  Abhishek Gupta \\
  Research Engineer \\
  University of Mumbai \\
  Mumbai, India\\
  \texttt{abhishekgupta@sjcem.edu.in} \\
    \And
  Raunak Joshi \\
  Mentor \\
  University of Mumbai \\
  Mumbai, India\\
  \texttt{raunakjoshi.m@gmail.com} \\
  \And
  Dr.~Vidya Chitre \\
  Assistant Professor \\
  Vidyalankar Institute of Technology \\
  Mumbai, India \\
  \texttt{vidya.chitre@vit.edu.in}
}

\renewcommand{\shorttitle}{Combining Varied Learners for Binary Classification using Stacked Generalization}

\hypersetup{
pdftitle={Combining Varied Learners for Binary Classification using Stacked Generalization},
pdfsubject={cs.LG},
pdfauthor={Sruthi Nair, Abhishek Gupta, Raunak Joshi, Vidya Chitre},
pdfkeywords={Ensemble Learning, Generalizing Error, Stacked Generalization},
}

\begin{document}
\maketitle

\begin{abstract}
	The Machine Learning has various learning algorithms that are better in some or the other aspect when compared with each other but a common error that all algorithms will suffer from is training data with very high dimensional feature set. This usually ends up algorithms into generalization error that deplete the performance. This can be solved using an Ensemble Learning method known as Stacking commonly termed as Stacked Generalization. In this paper we perform binary classification using Stacked Generalization on high dimensional Polycystic Ovary Syndrome dataset and prove the point that model becomes generalized and metrics improve significantly. The various metrics are given in this paper that also point out a subtle transgression found with Receiver Operating Characteristic Curve that was proved to be incorrect.
\end{abstract}

\keywords{Ensemble Learning \and Generalizing Error \and Stacked Generalization}

\section{Introduction}
Deriving inference from a very high dimensional data for classification tasks is very difficult for Machine Learning \citep{10.5120/ijca2017913083} techniques no matter how good they are operationally. Inference that is not able to derive is termed as error for machine learning models and results in loss \citep{yessou2020comparative} of information. Now there are techniques known as loss optimizers \citep{8959681} for deep learning \citep{lecun2015deep} techniques, but machine learning techniques require some for generalization of errors \citep{jakubovitz2019generalization}. If we consider machine learning, the wide variety of algorithms are available in parametric as well as non-parametric learning methods. The learning methods are about how efficiently they fit the data no matter how high are the dimensions. If we consider classification, we will specifically give importance to binary classification \citep{9230877} as it is the type of classification we are considering for this paper. Logistic Regression \citep{doi:10.1080/00220670209598786} is one of the most commonly used binary classification algorithm. The Logistic Regression is parametric and supervised learning algorithm and uses a logit function that gives 2 separate classes effectively. Even after working with hyperparameters of the algorithm, there can be many algorithms that work efficiently in different aspects of data. The extension of linear models can be algorithms like K-Nearest Neighbors, Support Vector Machines and many more. Right before diving into algorithms and techniques for generalizing errors in machine learning, let us consider the data we have used. Our selection criteria for the data was not limited as we wanted a data with very high number of dimensions with categorical variables and binary classification problem. The data that we considered for this paper is Polycystic Ovary Syndrome \citep{azziz2016polycystic} Classification. This data is specifically binary classification problem which based on the features gives the presence of symptom. The data is very high dimensional in terms of features and there are many categorical features. In order to harness any machine learning algorithm performance, categorical variables are challenging as the basis of it when converted to numerical depends on its purpose. The quantitative variables are very hard distinguish for algorithms and that is where the test of its performance starts. Conversion of such variables is necessary into numerical formats and it depends on the fact if they need to be converted into rank oriented or occurrence oriented. There are many facets for such kind of data and that is where we decided to work with it. If we consider polycystic ovary syndrome for machine learning, the problem is tackled by using logistic regression \citep{Li2011ALR}, bagging ensemble \citep{kanvinde2022binary}, discriminant analysis \citep{gupta2022discriminant}, boosting ensemble \citep{9697163} methods in these papers. This paper primarily focuses on techniques to generalize the errors of other models and secondarily focuses on application of the methodology given in this paper with polycystic ovary syndrome data.

\section{Methodology}
This section of paper specifically gives detailed procedure of how we approached the problem. In order to work with generalization error, we will use Stacking Ensemble Method which is also known as Stacked Generalization \citep{WOLPERT1992241,Ting97stackedgeneralization:}. First we will thoroughly walk you through all the classifiers used in stack of stacked generalization and then give detailed explanation of the stacked model.

\subsection{Logistic Regression}
This is going to be one of the algorithm that will contribute to the stack. The logistic regression is parametric supervised linear learning method. The logistic regression calculates the fixed parameters and uses them to calculate the prediction equation which is very similar to the linear regression. The formula for prediction function for calculating one single feature is given as

\begin{equation}
    \hat{y} = \gamma_0 + \gamma_i.(x_i)
\end{equation}

The $\gamma$ is just the arbitrary form of variable and any other notation can be used. This equation is for fitting a linear line over the data points, but since logistic regression is classification algorithm, the line needs to be converted with logit function to give distinct separation between the classes. This function for logit is given as

\begin{equation}
    \sigma(Z) = \frac{1}{1+e^{-Z}}
\end{equation}

where $\sigma$ is termed as the logit function also known as Sigmoid function and $Z$ is basically linear function.

\subsection{Support Vector Machine}
The support vector machine \citep{708428,Evgeniou2001SupportVM,cristianini2008support} abbreviated as SVM is a non-parametric supervised learning algorithm that uses the hyperplane function to estimate the points in the data. The hyperplane is a line that is passed between different data points for differentiating classes separately. The hyperplane is a single line and requires maximal margin that to get closer to points. The points that touch the margin are known as Support Vectors. The points that first touch the maximal margin is the classified label. The hyperplane calculation requires vector normal and an offset point. This can be represented in equation as

\begin{equation}
    w^T.x+b=0
\end{equation}

where $w$ is the vector normal and $b$ is the offset point. The maximal margin is later given with +/- equation and can be represented as

\begin{equation}
    \begin{cases}
      +, & w^T.x+b>0 \\
      -, & w^T.x+b<0
    \end{cases}
\end{equation}

\subsection{Multi Layer Perceptron}
The term of Multi Layer Perceptron \citep{Ramchoun2016MultilayerPA} abbreviated as MLP is coined from the field of Deep Learning. The field of deep learning is very widely used now and then but the foundation of it is Perceptron. The term of perceptron was coined for the fact that system was able to develop human brain abilities that included perception. The perceptron works with initially, initialized weights and biases that influence the learning process of the representations from the data. These weights and biases requires an activation function \citep{nwankpa2018activation} to retain the features and details of the data. Logit function of Logistic Regression is actually a type of activation function that is used for binary classification. Similarly for learning information in a greater amount the perceptron is used in a connection which is known as Multi Layer. The activation function in initial perceptron requires to learn features and can use different activation functions. The Rectified Linear Unit \citep{agarap2019deep} also known as ReLU is one of the non-linearity activation function.

\subsection{Random Forest}
The random forest \citep{Breiman2004RandomF} is non-parametric supervised bagging ensemble learning method. Since it is an ensemble learning method it actually consists of weak learners. It was actually developed to overcome the drawbacks of Decision Tree \citep{10.1023/A:1022643204877} which suffer from high variance \citep{Geurts2001ImprovingTB} problem when the number of features are very high. The random forest is also considered as bootstrap aggregation method as the process includes combining a set of trees and considering their final outcome. The accuracy of random forest turns out to be much better than many of algorithms but still there are chances it can suffer from generalization errors.

\subsection{K-Nearest Neighbors}
The K-Nearest Neighbors \citep{Cunningham2020kNearestNC} abbreviated as KNN is a non-parametric supervised distance based learning method. The algorithm uses distance learning equations for estimation of classes with the points newly available for prediction. The majority of votes is taken for prediction classes based on highest probabilities. The distance functions used can be Euclidean, Minkowski and Manhattan distances.

\subsection{Stacked Generalization}
This is where the important part of the paper lies, the stacked generalization \citep{WOLPERT1992241,Ting97stackedgeneralization:} model and we will explain the methodology of it along with the things discussed earlier. The stacked generalization has 2 phases, first phase the stack of diversified models are trained and their predictions are stored. Second phase uses the stored predictions from the first phase as an input and trains the model. This phase is also known as meta-classifier and its output is the final prediction made for the data. The depiction can be done in the Figure \ref{fig:fig1} given below.

\begin{figure}[htbp]
	\centering
	\includegraphics{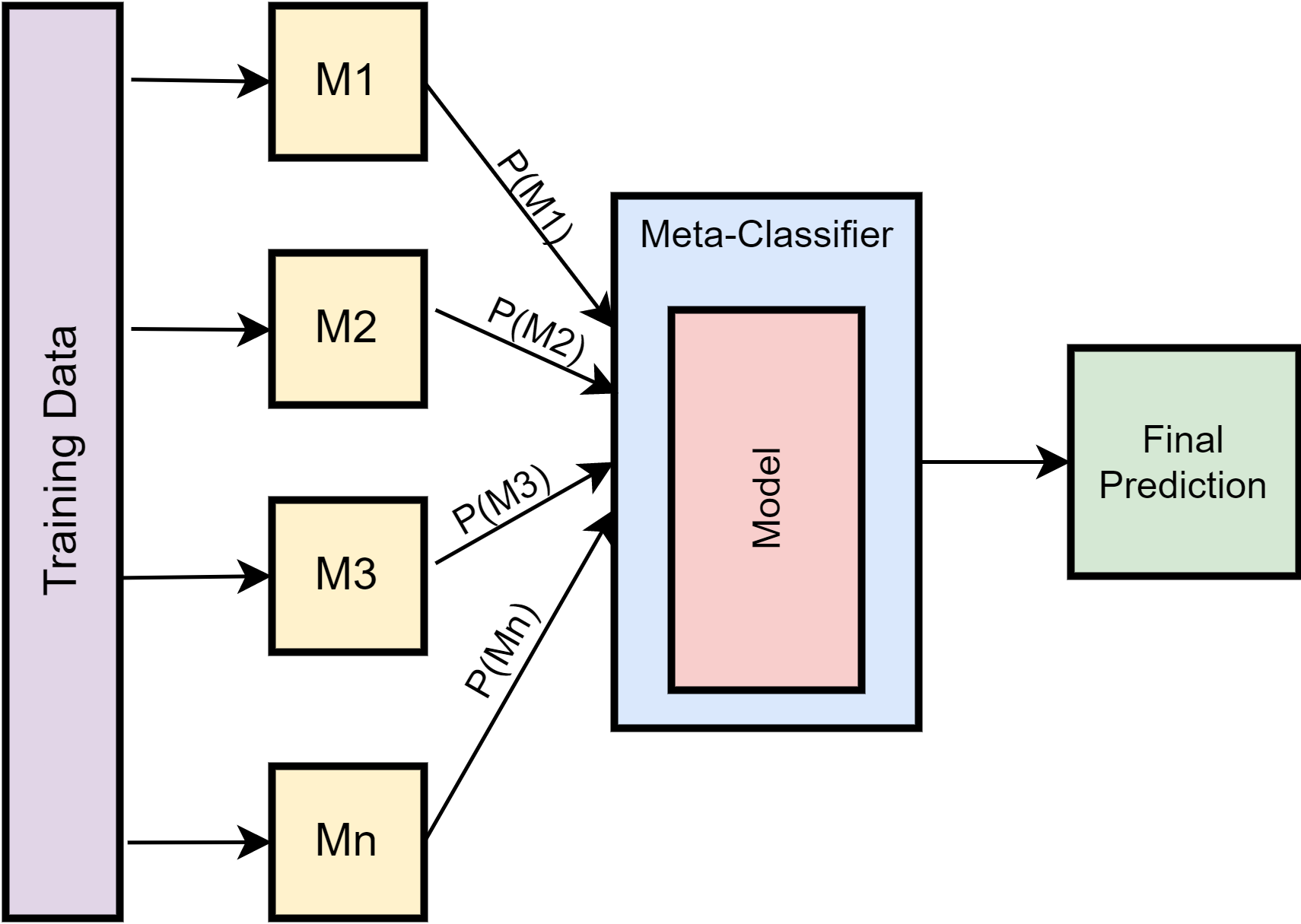}
	\caption{Stacked Generalization}
	\label{fig:fig1}
\end{figure}

The models in first phase of the Figure \ref{fig:fig1} are independent from each other. These can be parametric and non-parametric, can be very high in numbers and can be combination of same models with different hyperparameters. The predictions of all the models are yielded and later given to the meta-classifier block. This meta-classifier is a model itself that can be parametric or non-parametric in nature and uses the predictions of the models from the first phase as an input as shown in the figure. This later gives the final output. This process generalizes the errors of the model and works to overcome the limitations of the models used in an independent manner. For the implementation the models that we have used are support vector machine, random forest, multi-layer perceptron and k-nearest neighbors. The predictions of these models are later given to meta-classifier for which we have used the linear model, logistic regression. This gives us final output.

\begin{figure}[htbp]
	\centering
	\includegraphics[scale=0.8]{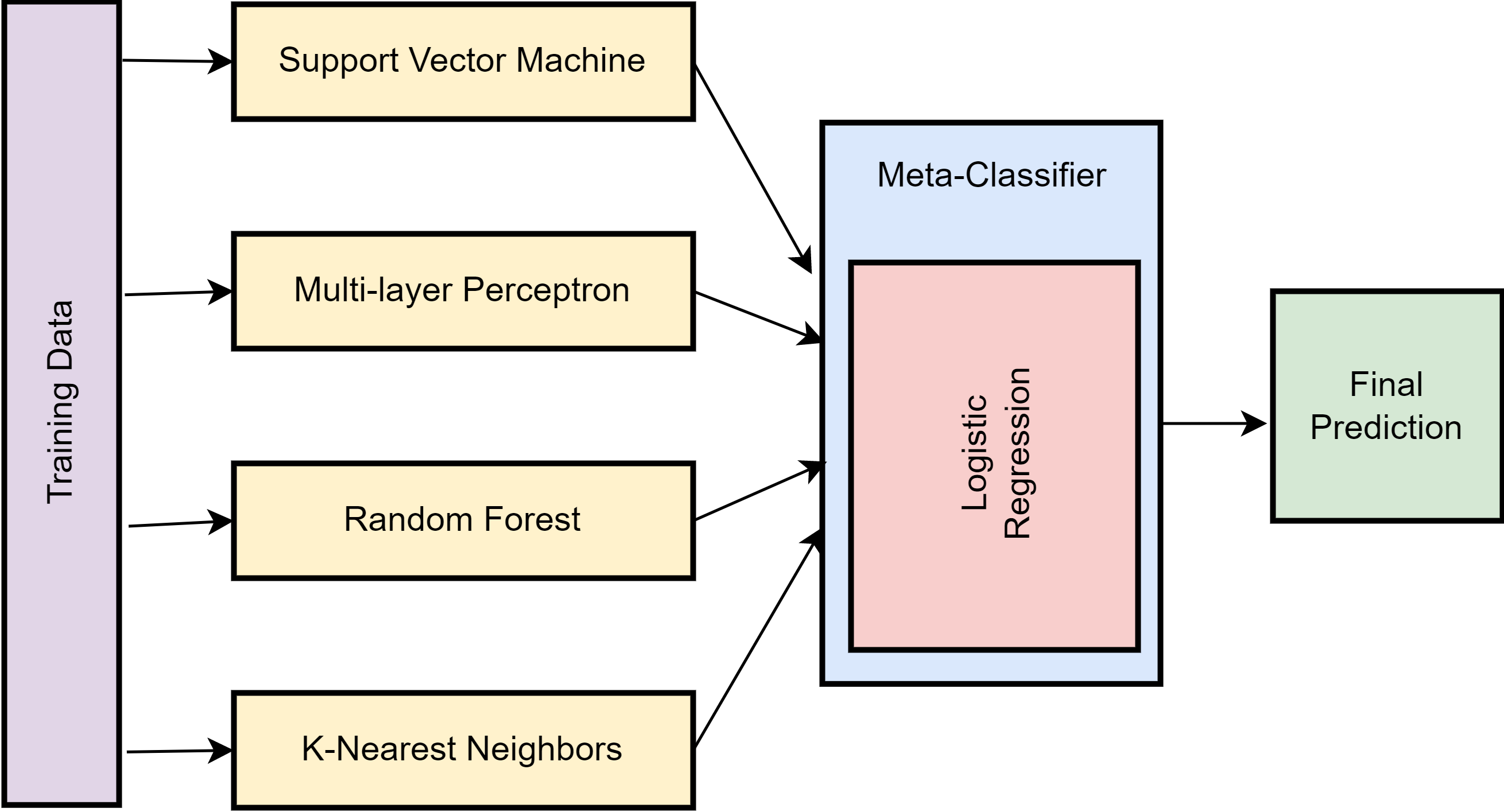}
	\caption{Stacked Generalization Models Used}
	\label{fig:fig2}
\end{figure}

The Figure \ref{fig:fig2} is the model of our implementation, where 4 models are used in the first phase of the stacking and generalization is done with meta-classifier using Logistic Regression. The support vector machine hyper-parameters used are Radial Basis Filtering \citep{5698379} kernel on third degree with gamma scaling. The hyper-parameters for Multi-layer perceptron are ReLU activation function, alpha values as 0.1 and hidden layers 1000. The Random Forest hyper-parameters are 500 estimators, criterion as gini index, maximum depth as 10 and minimum leaf samples with 0.005 value. The K-Nearest Neighbor value for k is 5 and weights selected are uniform. The stacking uses Logistic Regression with default hyper-parameters and cross validation with 5 folds.

\section{Result}
This section gives a detailed analysis of the outcomes generated by the implementation. There are many methods for deriving the metrics of the model.

\subsection{Precision and Recall}
The precision and recall \citep{powers2020evaluation} are the very first of the metrics that are derived for basic inference of the models. Obviously those are not the only metrics to be considered for final evaluation but they acts as the preliminary metrics for further methods. The precision and recall is considered with macro and weighted average.

\begin{table}[htbp]
	\caption{Precision and Recall}
	\centering
	\begin{tabular}{lllll}
		\toprule
		Algorithm & Macro Precision & Weighted Precision & Macro Recall & Weighted Recall \\
		\midrule
		SVM & 88\% & 87\% & 82\% & 85\% \\
		MLP & 78\% & 78\% & 75\% & 78\% \\
		RF & 88\% & 87\% & 82\% & 85\% \\
		KNN & 87\% & 85\% & 79\% & 84\% \\
		SG & 91\% & 89\% & 83\% & 87\% \\
		\bottomrule
	\end{tabular}
	\label{tab:tab1}
\end{table}

The Table \ref{tab:tab1} gives a detailed analysis of Precision and Recall and Stacked Generalization is giving a better result as compared to other algorithms. But these metrics are not enough to judge the performance of algorithms.

\subsection{F Measures}
The harmonic average for entire model can be calculated using combination of Precision and Recall using the F Measures. There are many variations in the f measure. The formula for basic f measure can be given as

\begin{equation}
    F = 2.\frac{Precision . Recall}{Precision + Recall}
\end{equation}

The result value for F Measure which is also known as F-Score \citep{powers2020evaluation} is given between 0 and 1. Closer the value to 1, better the accuracy of model and vice versa. Table \ref{tab:tab2} gives a context for F Measures in different averages.

\begin{table}[htbp]
	\caption{F Measure in different Averages}
	\centering
	\begin{tabular}{llll}
		\toprule
		Algorithm & Accuracy F-Score & Macro F-Score & Weighted F-Score \\
		\midrule
		SVM & 85\% & 83\% & 85\% \\
		MLP & 78\% & 76\% & 78\% \\
		RF & 85\% & 83\% & 85\% \\
		KNN & 84\% & 81\% & 83\% \\
		SG & 87\% & 85\% & 87\% \\
		\bottomrule
	\end{tabular}
	\label{tab:tab2}
\end{table}

Before confirming the inference directly based on F-Score, one different angle can be given to the F-Score that control over the balancing Precision and Recall where $\beta$ parameter is used. This type of variation in F-Score is $F_\beta$-Score where $\beta$ is the coefficient for balance. The formula for $F_\beta$-Score can be given as

\begin{equation}
    F_\beta = \frac{(1+\beta^2).P.R}{\beta^2.(P+R)}
\end{equation}

The $\beta$ values usually considered are 0.5, 1 and 2. The $\beta$ parameter as 0.5 indicates more weight on precision and less weight on recall. The $\beta$ parameter as 1 indicates the balanced precision and recall which is very similar to traditional F measure. Selecting the $\beta$ parameter as 2 indicates less weight on precision and more weight on recall. We can see the values with the $\beta$ parameter in the Table \ref{tab:tab3} given below.

\begin{table}[htbp]
	\caption{$F_\beta$ with 0.5, 1 and 2 as $\beta$ value}
	\centering
	\begin{tabular}{llll}
		\toprule
		Algorithm & $F_{0.5}$-Score & $F_{1}$-Score & $F_{2}$-Score \\
		\midrule
        SVM & 85.73\% & 83.48\% & 82.22\% \\
        MLP & 76.83\% & 75.87\% & 75.28\% \\
        RF & 85.73\% & 83.48\% & 82.22\% \\
        KNN & 83.87\% & 81.14\% & 79.77\% \\
        SG & 88.38\% & 85.33\% & 83.73\% \\
		\bottomrule
	\end{tabular}
	\label{tab:tab3}
\end{table}

The higher generalization criterion gives a better intuition of the learning patterns of the model. This is the factor that clearly influences the model. This improvement in score is not because of better leverage of widespread in algorithm but a better measure of generalization of the model which is not achieved from independent models. The widespread of the algorithm can be seen with Receiver Operator Characteristics \citep{Bradley1997TheUO} and Area Under Curve. The RoC score to some extent of Random Forest is better than Stacked Generalization yet the F measure metrics give a completely different observation. This can be visualized efficiently in in Figure \ref{fig:fig3} given below.

\begin{figure}[htbp]
    \centering
    \includegraphics[scale=0.57]{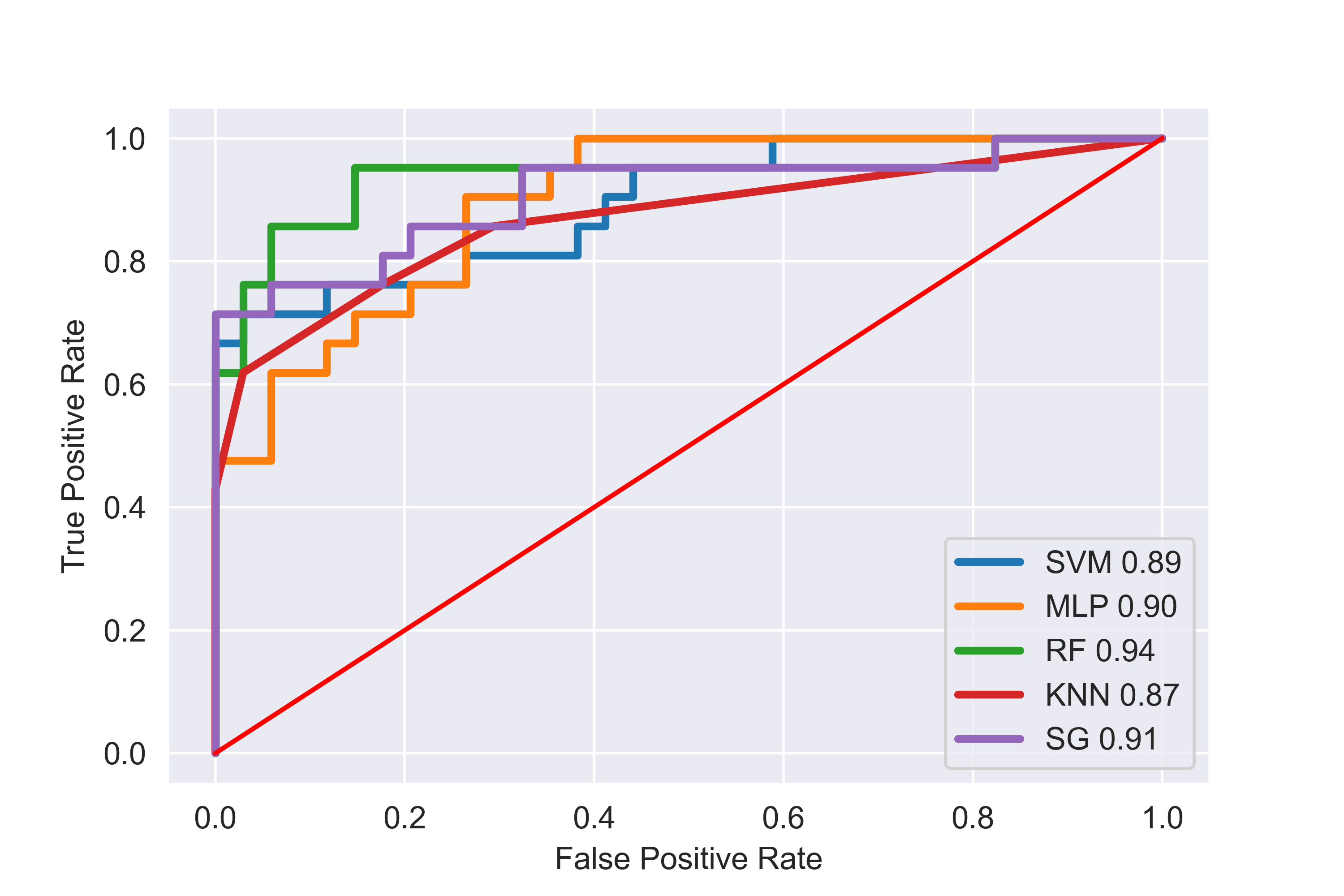}
    \caption{RoC and AUC}
    \label{fig:fig3}
\end{figure}

\begin{figure}[hbt]
  \centering
  \begin{tabular}{ll}
  \includegraphics[scale=0.37]{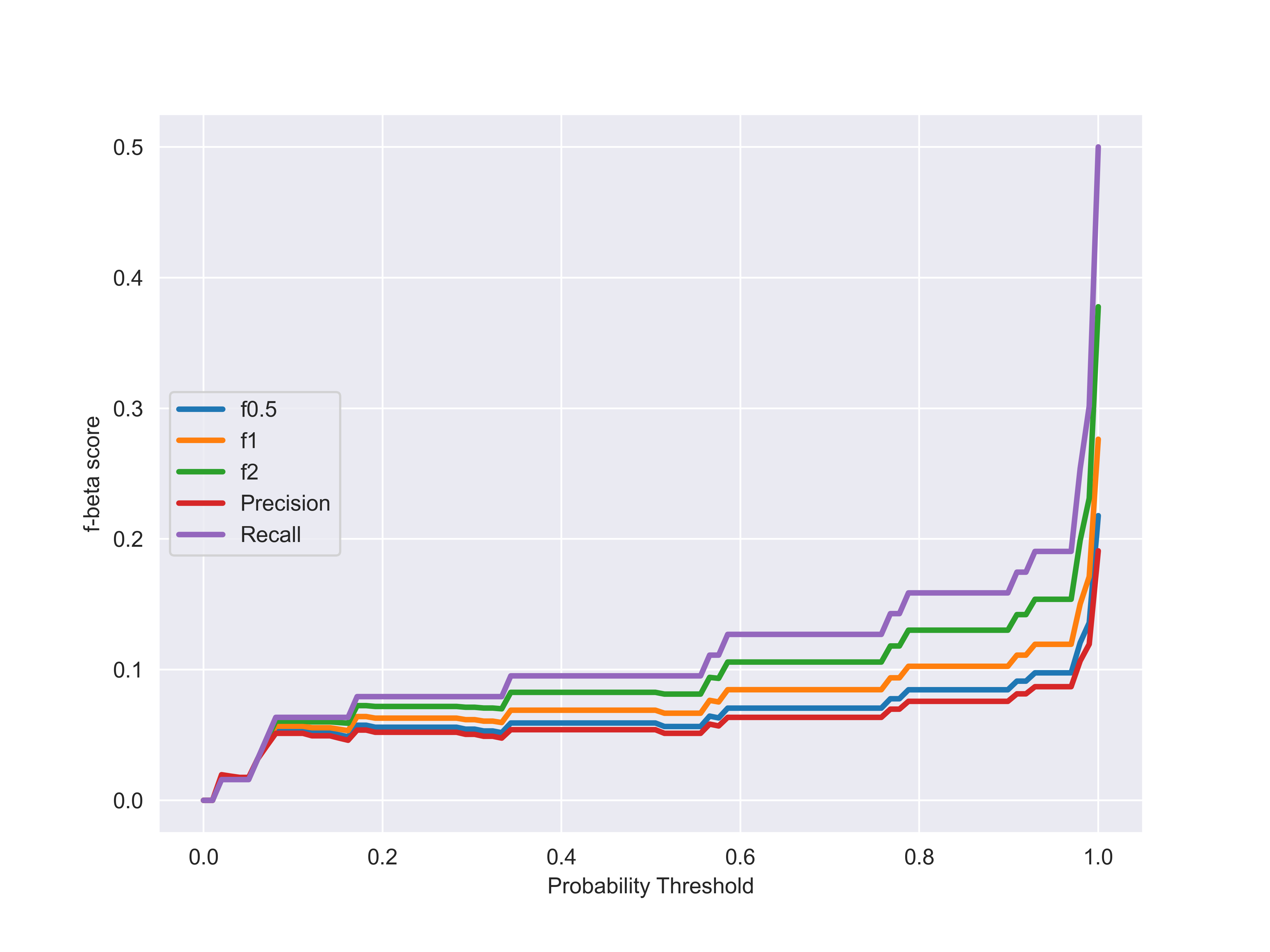}
  &
  \includegraphics[scale=0.37]{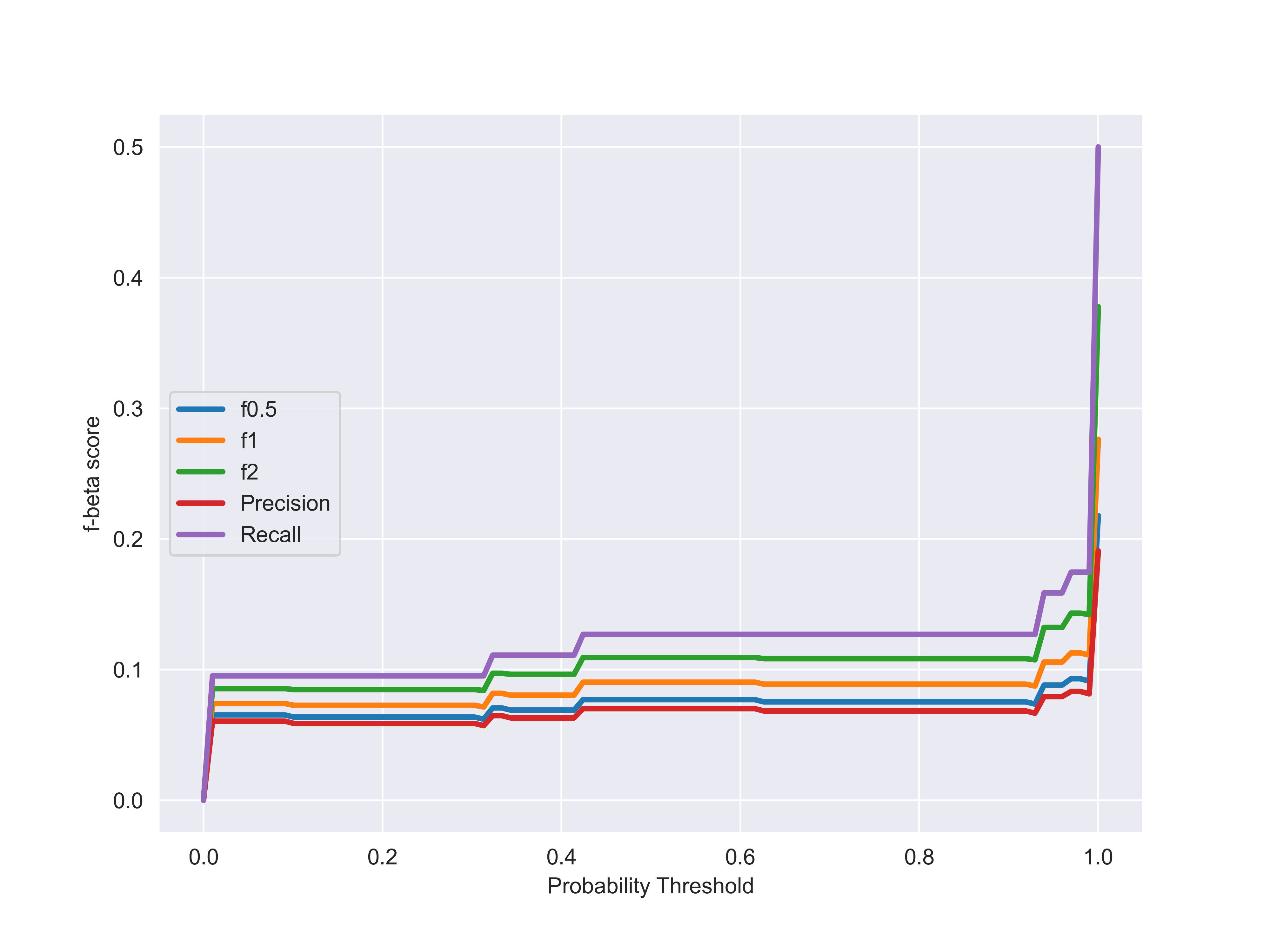}
  \end{tabular}
  \begin{tabular}{ll}
  \includegraphics[scale=0.37]{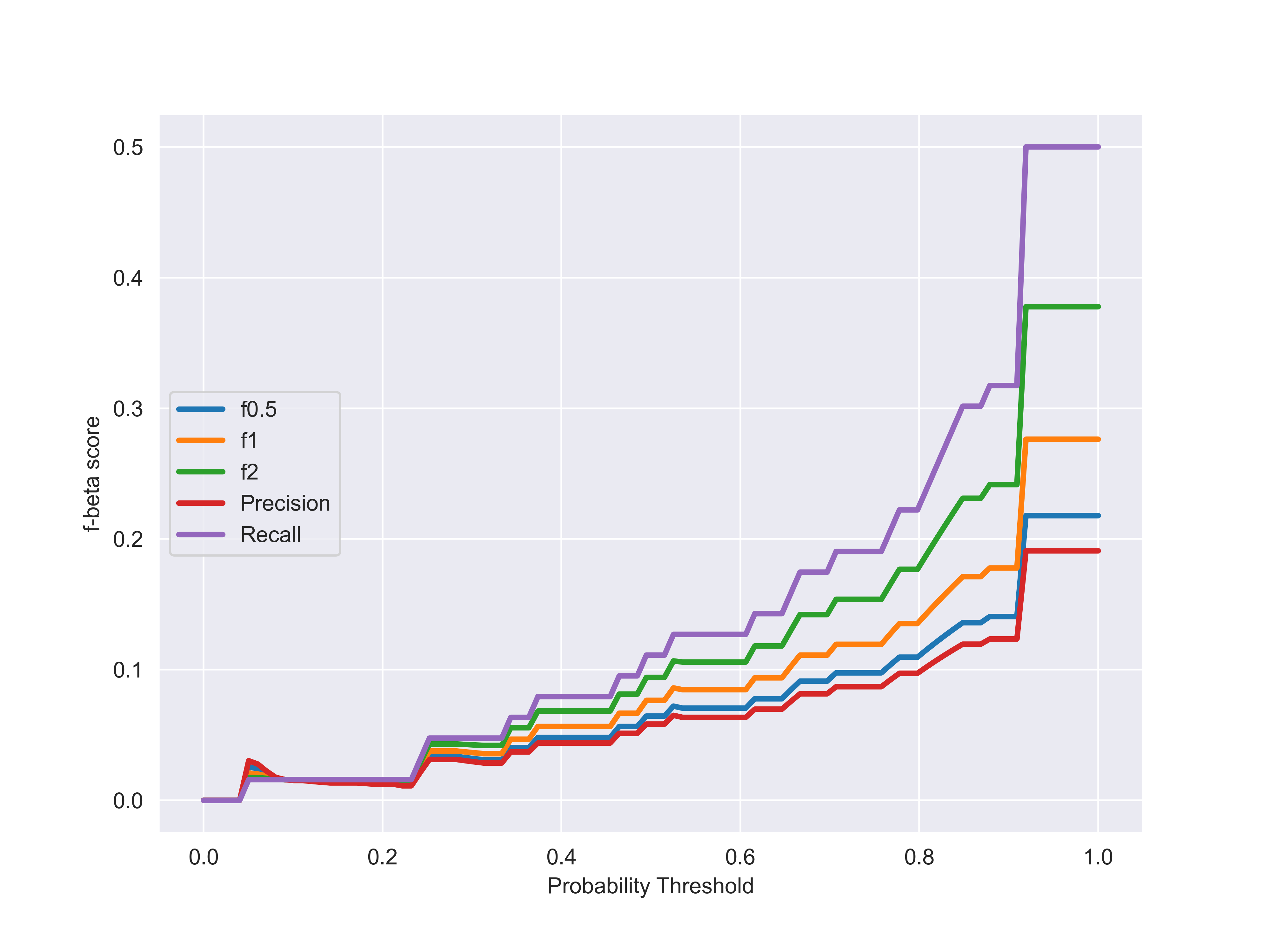}
  &
  \includegraphics[scale=0.37]{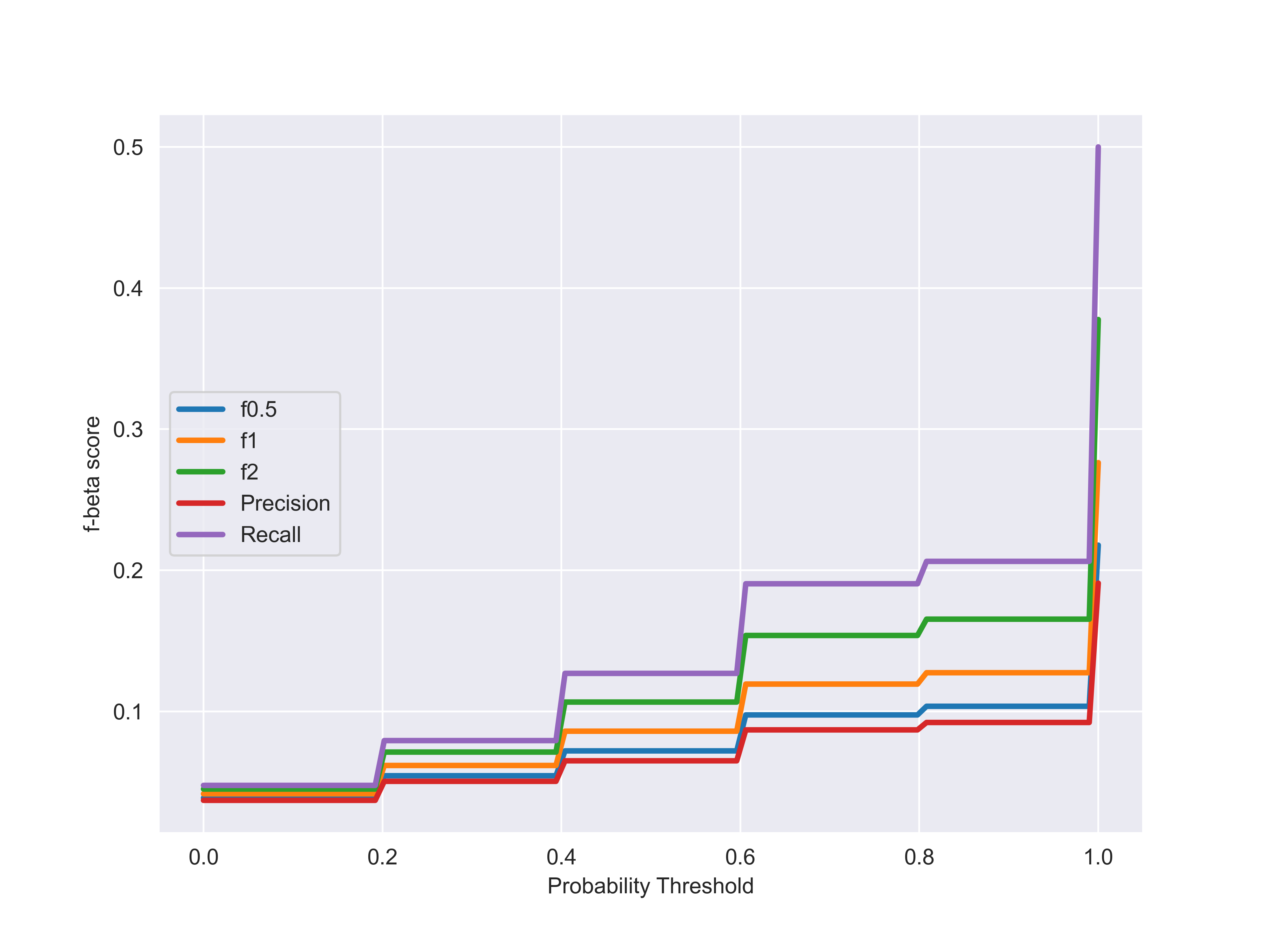}
  \end{tabular}
  \begin{tabular}{l}
  \includegraphics[scale=0.37]{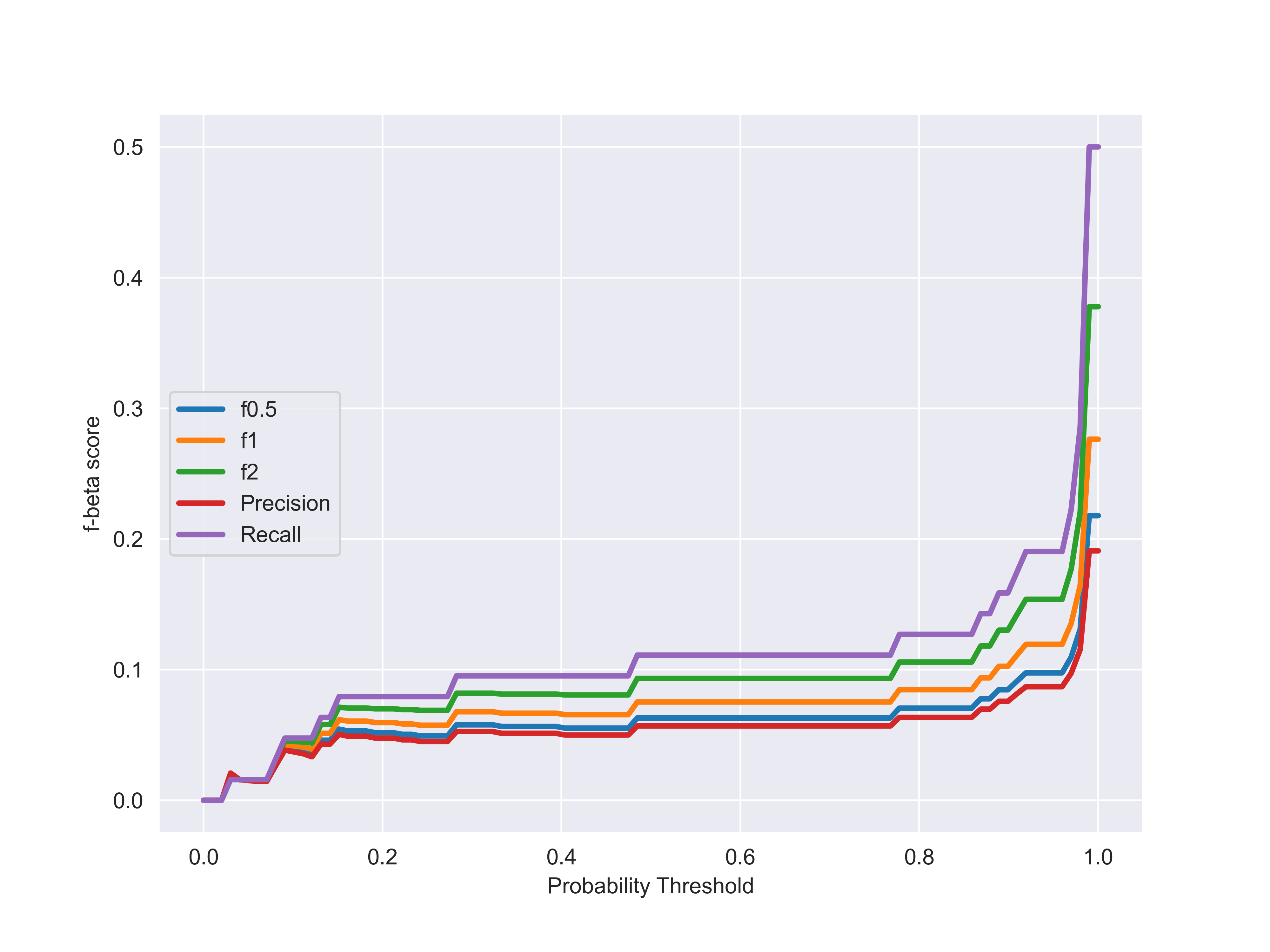}
  \end{tabular}
  \caption{F Scores in multiple contexts for SVM, MLP, RF, KNN and Stacked}
  \label{fig:fig4}
\end{figure}

For this we will use other metrics too but we can also try visualizing the f scores with precision and recall in Figure \ref{fig:fig4}.

\subsection{Hamming Loss}
Just to ensure and get rid of the confusion of performance, we have some other metrics too. Hamming Loss \citep{Tsoumakas07multi-labelclassification:} is one of them which indicates the wrong labels fraction with respect to total number of labels. The formula can be represented as

\begin{equation}
    L_{Hamming} = \frac{1}{NL} \sum_{i=1}^N\sum_{l=1}^L \oplus(Y_{i,l}, \hat{Y}_{i,l})
\end{equation}

The N is number of labels, L is density of labels, Y is true labels and $\hat{Y}$ is predicted labels. The score of Hamming Loss is always between 0 and 1. Closer to 0, lesser the loss which indicates better model and with 1 is vice versa. The scores can be seen in the form of table.

\begin{table}[htbp]
	\caption{Hamming Loss}
	\centering
	\begin{tabular}{ll}
		\toprule
		Algorithm & Hamming Loss \\
		\midrule
        SVM & 14.54\% \\
        MLP & 21.81\% \\
        RF & 14.54\% \\
        KNN & 16.36\% \\
        SG & 12.72\% \\
		\bottomrule
	\end{tabular}
	\label{tab:tab4}
\end{table}

The Table \ref{tab:tab4} indicates the Stacked Generalization is performing the best as it has the value closest to zero. This pretty much sums up the quality of stacked generalization, but one more metric can be used to prove the point.

\subsection{Jaccard Index}
The Jaccard \citep{6268901} is an accuracy metric that uses set theory principles. The formula for it is given by

\begin{equation}
    I_{jaccard}(Y, \hat{Y}) = \frac{|Y \bigcap \hat{Y}|}{|Y \bigcup \hat{Y}|}
\end{equation}

The I is jaccard index, Y is true labels and $\hat{Y}$ is predicted labels. The intersection and union are used to evaluate the score. Closer to 1, better the accuracy and with 0, vice versa. The jaccard index for the algorithms is given below in Table \ref{tab:tab5}.

\begin{table}[htbp]
	\caption{Jaccard Index}
	\centering
	\begin{tabular}{ll}
		\toprule
		Algorithm & Jaccard Index \\
		\midrule
        SVM & 74.60\% \\
        MLP & 64.17\% \\
        RF & 74.60\% \\
        KNN & 71.87\% \\
        SG & 77.41\% \\
		\bottomrule
	\end{tabular}
	\label{tab:tab5}
\end{table}

The Jaccard Index for all the algorithms is given and Stacked Generalization performs the best as compared to all other algorithms. This is possible because of generalization ability of the model.

\section{Conclusion}
In this paper we aimed for a very subtle point in Machine Learning domain that is usually not given much preference. The algorithms learn representations in an independent fashion, have limitations when they learn from a very high dimensional data with a lot of categorical variables. When these categorical variables are preprocessed in numerical values, they might either create more features else increase the complexity of learning for algorithm. This higher complexity issue causes the algorithms to become less generalized and induce a lot of errors. For this, in this paper we presented a long lost topic of Stacked Generalization Ensemble learning that helps create less generalized models in machine learning without switching to deep learning. If the number of features is high and records are less, this is the best method to implement. We used PCOS classification example to highlight this point and provided a good comparison of Stacked Generalization with a lot of popularly used algorithms and proved the point that some metrics can be misleading and many diversified metrics should be used to yield best results. This paper definitely gives a lot of insights on varied topics and we would like to see someone contributing with this paper to their work with our best belief and knowledge.

\section*{Acknowledgment}
We appreciate the effort of Mr. Prasoon Kottarathil and sincerely thank him for providing Polycystic Ovary Synodrome dataset on Kaggle.

\bibliographystyle{unsrtnat}
\bibliography{references}
\end{document}